# Machine learning methods for histopathological image analysis

Daisuke Komura[1], Shumpei Ishikawa[1]*

* Corresponding author

[1]Department of Genomic Pathology, Medical Research Institute, Tokyo Medical and Dental University, Tokyo, Japan

Abstract

Abundant accumulation of digital histopathological images has led to the increased demand for their analysis, such as computer-aided diagnosis using machine learning techniques. However, digital pathological images and related tasks have some issues to be considered. In this mini-review, we introduce the application of digital pathological image analysis using machine learning algorithms, address some problems specific to such analysis, and propose possible solutions.

1. Introduction

Pathology diagnosis has been performed by a human pathologist observing the stained specimen on the slide glass using a microscope. In recent years, attempts have been made to capture the entire slide with a scanner and save it as a digital image (Whole slide image, WSI) [1]. As a large number of WSI are being accumulated, attempts have been made to analyze WSIs using digital image analysis based on machine learning algorithms to assist tasks including diagnosis.

Digital pathological image analysis often uses general image recognition technology (e.g. facial recognition) as a basis. However, since digital pathological images and tasks have some unique characteristics, special processing techniques are often required. In this review, we describe the application of digital pathological image analysis using machine learning algorithms, and its problems specific to digital pathological image analysis and the possible solutions. Several reviews that have been published recently discuss histopathological image analysis including its history and details of general machine learning algorithms [2–7]; in this review, we provide more pathology-oriented point of view.

Since the overwhelming victory of the team using deep learning at ImageNet Large Scale Visual Recognition Competition (ILSVRC) 2012 [8], most of the image recognition techniques have been replaced by deep learning. This is also true for pathological image analysis [9–11]. Therefore, even though many techniques introduced in this review are related to deep learning, most of them are also applicable for other machine learning algorithms.

2. Machine learning methods

Figure 1 shows typical steps for histopathological image analysis using machine learning. Prior to applying machine learning algorithms, some pre-processing should be performed. For example, when cancer regions are detected in WSI, local mini patches around 256 × 256 are sampled from large WSI. Then feature extraction and classification between cancer and non-cancer are performed in each local patch. The goal of feature extraction is to extract useful information for machine learning tasks. Various local features such as gray level co-occurrence Matrix (GLCM) and local binary pattern (LBP) have been used for histopathological image analysis, but deep learning algorithms such as convolutional neural network [9,10,12–14] starts the analysis from feature extraction. Features and classifiers are simultaneously optimized in deep learning and features learned in deep learning often outperforms other traditional features in histopathological image analysis.

Machine learning techniques often used in digital pathology image analysis are divided into supervised learning and unsupervised learning. The goal of supervised learning is to infer a function that can map the input images to their appropriate labels (e.g. cancer) well using training data. Labels are associated with a WSI or an object in WSIs. The algorithms for supervised learning include support vector machines, random forest and convolutional neural networks. On the other hand, the goal of unsupervised learning is to infer a function that can describe hidden structures from unlabeled images. The tasks include clustering, anomaly detection and dimensionality reduction. The algorithms for unsupervised learning include k-means, autoencoders and principal component analysis. There are derivatives from these two learning such as semi-supervised learning and multiple instance learning, which are described in Section 4.2.2.

3. Machine learning application in digital pathology

3.1. Computer-assisted diagnosis

The most actively researched task in digital pathological image analysis is computer-assisted diagnosis (CAD), which is the basic task of the pathologist. Diagnostic process contains the task to map a WSI or multiple WSIs to one of the disease categories, meaning that it is essentially a supervised learning task. Since the errors made by a machine learning system reportedly differ from those made by a human pathologist [15], classification accuracy could be improved using CAD system. CAD may also lead to the reduce variability in interpretations and prevent overlooking by investigating all pixels within WSIs.

Other diagnosis-related tasks include detection or segmentation of Region of Interest (ROI) such as tumor region in WSI [16,17], scoring of immunostaining [11,18], cancer staging [15,19], mitosis detection [20,21], gland segmentation [22–24], and detection and quantification of vascular invasion [25].

3.2. Content Based Image Retrieval

Content Based Image Retrieval (CBIR) retrieves similar images to a query image. In digital pathology, CBIR systems are useful in many situations, particularly in diagnosis, education, and research [26–31]. For example, CBIR systems can be used for educational purposes by students and beginner pathologists to retrieve relevant cases or histopathological images of tissues. In addition, such systems are also helpful to professional pathologists, particularly when diagnosing of rare cases.

Since CBIR does not necessarily require label information, unsupervised learning can be used [30]. When label information is available, supervised learning approaches could learn better similarity measure than unsupervised learning approaches [28,29] since the similarity between histopathological images may differ by definition. However, preparing sufficient number of labeled data can be a serious problem as will be described later.

In CBIR, not only accuracy but also high-speed search of similar images from numerous images are required.

Therefore, various techniques for dimensionality reduction of image features such as principal component analysis and compact bilinear pooling [32], and fast approximate nearest neighbor search such as kd-tree and hashing [33] are utilized for high speed search.

3.3. Discovering new clinicopathological relationships

Historically, many important discoveries concerning diseases such as tumor and infectious diseases have been made by pathologists and researchers who have carefully and closely observed pathological specimens. For example, H. pylori was discovered by a pathologist who was examining the gastric mucosa of patients with gastritis [34]. Attempts have also been made to correlate the morphological features of cancers with their clinical behavior. For example, tumor grading is important in planning treatment and determining a patient's prognosis for certain types of cancer, such as soft tissue sarcoma, primary brain tumors, and breast and prostate cancer.

Meanwhile, thanks to the progress in digitization of medical information and advance in genome analysis technology in recent years, large amount of digital information such as genome information, digital pathological images, MRI and CT images has become available [35]. By analyzing the relationship between these data, new clinicopathological relationships, for example, the relationship between the morphological characteristic and the somatic mutation of the cancer, can be found [35,36]. However, since the amount of data is enormous, it is not realistic for pathologists and researchers to analyze all the relationships manually by looking at the specimens. This is where the machine learning technology comes in. For example, Beck et al. extracted texture information from pathological images of breast cancer and analyzed with L1 - regularized logistic regression, and indicated that the histology of stroma correlates with prognosis in breast cancer [37]. Other researches include prognosis predictions from histopathological image of cancer [38], prediction of somatic mutation [13], and discovery of new gene variants related to autoimmune thyroiditis based on image QTL [39].

4. Problems specific to histopathological image analysis

In this section, we describe unique characteristics of pathological image analysis and computational methods to treat them. Table 1 presents an overview of papers dealing with the problems and the solutions.

Table 1. Overview of papers dealing with problems and solutions for histopathological image analysis

| Solution | reference |
|---|---|
| **Very large image size** | |
| Case level classification summarizing patch or object level classification | Markov Random Field [17], Bag of Words of local structure [18] and random forest [14,40,41] |
| **Insufficient labeled images** | |
| GUI tools | Web server [42,43] |
| Tracking pathologists' behavior | Eye tracking [44], mouse tracking [45] and viewport tracking [46] |
| Active learning | Uncertainly sampling [43], Query-by-Committee [47], variance reduction [48] and hypothesis space reduction [49] |
| Multiple instance learning | Boosting-based [50,51], deep weak supervision [52] and structured support vector machines (SVM) [53] |
| Semi-supervised learning | Manifold learning [30] and SVM [54] |
| Transfer learning | Feature extraction [55], fine-tuning [16,56,57] |
| **Different levels of magnification result in different levels of information** | |
| Multiscale analysis | CNN [58], dictionary learning [59] and texture features [60] |
| **WSI as orderless texture-like image** | |
| Texture features | Traditional textures [61–64] and CNN-based textures [65] |
| **Color variation and artifacts** | |
| Removal of color variation effect | Color normalization [66–69] and color augmentation 70,71] |
| Artifact detection | Blur [72,73] and tissue-folds [74,75] |

## 4.1. Very large image size

When images such as dogs or houses are classified using deep learning, small sized image such as 256 × 256 pixels is often used as an input. Images with large size often need to be resized into smaller size which is enough for sufficient distinction, as increase in the size of the input image results in the increase in the parameter to be estimated, the required computational power, and memory. In contrast, WSI contains many cells and the image could consist of as many as tens of billions of pixels, which is usually hard to analyzed as is. However, resizing the entire image to a smaller size such as 256 × 256 would lead to the loss of information at cellular level, resulting in marked decrease of the identification accuracy. Therefore, the entire WSI is commonly divided into partial regions of about 256 × 256 pixels ("patches"), and each patch is analyzed independently, such as detection of ROIs. Thanks to the advances in computational power and memory, patch size is increasing (e.g. 960 × 960), which is expected to contribute to better accuracy. There is still a room for improvement in the method of integrating the result from each patch. For example, as the entire WSI could contain hundreds of thousands of patches, false positives are highly likely to appear even if individual patches are accurately classified. One possible solution for this is regional averaging of each decision, such that the regions is classified as ROI only when the ROI extends over multiple patches. However, this approach may suffer from false negatives, resulting in missing small ROIs such as isolated tumor cells [41].

In some applications such as IHC scoring, staging of lymph node metastasis of specimens or patients, and staging of prostate cancer diagnosed by Glisson score of multiple regions within one slide, more sophisticated algorithms to integrate patch-level or object-level decisions are required [14,17,18,40,41,76]. For example, for pN-staging of metastatic breast cancer, which was one of the tasks in Camelyon 17, multiple participating teams including us applied random forest classifiers of pixel or patch-level probabilities estimated by deep learning using various features such as estimated tumor size [41].

4.2. Insufficient labeled images

Probably the biggest problem in pathological image analysis using machine learning is that only a small number of training data is available. A key to the success of deep learning in general image recognition task is that training data is extremely abundant. Although label information at patch-level or pixel-level (e.g. inside/outside boundary of cancerous regions) is required in most tasks in digital pathology such as diagnosis, most labels of WSIs are at case-level (e.g. diagnosis) at most. Label information in general image analysis can be easily retrieved from the internet and it is also possible to use crowdsource labeling because anyone can identify objects and perform labeled work. However, only pathologists can label the pathological image accurately, and labeling at the regional level in a huge WSIs requires a lot of labor.

It is possible to reuse public ready-to-analyze data as training data in machine learning, such as ImageNet [77] in natural images and International Skin Imaging Collaboration [78] in macroscopic diagnosis of skin. In the field of digital pathology, there are some public datasets that contain hand-annotated histopathological images as summarized in Table 1 and Table 2. They could be useful if the purpose of the analysis, slide condition (e.g. stain), and image condition (e.g. magnification level and image resolution) are similar. However, because each of these datasets focuses on specific disease or cell types, many tasks are not covered by these datasets. There are also several large-scale histopathological image databases that contain high-resolution WSIs: The Cancer Genome Atlas (TCGA) [79] contains over 10000 WSIs from various cancer types, and Genotype-Tissue Expression (GTEx) [80,81]contains over 20000 WSIs from various tissues. These databases may serve as potential training data for various tasks. Furthermore, both TCGA and GTEx also provide genomic profiles, which could be used to investigate relationships between genotype and morphology. The problem is that the WSIs in these repositories contain labels at the case-level, and in order to be able to use them for training data, some preprocessing or specialized machine learning algorithm for treating case-level labels is required.

Many researches have attempted to solve the problem. Most of the approaches fall into one of the following categories: 1) efficient increase of label data, 2) utilization of weak label or unlabeled information, or 3) utilization of models/parameters for other tasks.

Table 2. downloadable WSI database.

| Dataset or author's name | # slides or patches | Stain | disease | Additional data |
|---|---|---|---|---|
| TCGA [34,82] | 18462 | H&E | cancer | genome/transcriptome/epigenome |
| GTEx [80,81] | 25380 | H&E | normal | transcriptome |
| TMAD [83,84] | 3726 | H&E / IHC | | IHC score |
| TUPAC16 [85] | 821 from TCGA | H&E | breast cancer | Proliferation score for 500 WSIs, position for mitosis for 73 WSIs, ROI for 148 cases |
| Camelyon17 [41] | 1000 | H&E | breast cancer (lymph node metastasis) | mask for cancer region (in 500 WSIs with 5 WSIs per patient) |
| Köbel et al. [53,86] | 80 | H&E | Ovarian carcinoma | |
| KIMIA Path24 [87,88] | 24 | H&E / IHC and others | various tissue | |

Table 3. Hand annotated histopathological images publicly available.

| Dataset or Paper | Image size(px) | # images | Stain | Disease | Additional data | Potential usage |
|---|---|---|---|---|---|---|
| KIMIA960 [89,90] | 308x168 | 960 | H&E / IHC | various tissue | | disease classification |
| Bio-Segmentation [91,92] | 896x768, 768x512 | 58 | H&E | Breast cancer | | disease classification |
| Bioimaging Challenge 2015[93,94] | 2040x1536 | 269 | H&E | Breast cancer | | disease classification |
| GlaS [23,95] | 574-775 x 430-522 | 165 | H&E | Colorectal cancer | mask for gland area | gland segmentation |
| BreakHis [15,96] | 700 x 460 | 7909 | H&E | Breast cancer | | disease classification |
| Jakob Nikolas et al. [90,97] | 1000 x 1000 | 100 | IHC | colorectal cancer | blood vessel count | blood vessel detection |
| MITOS-ATYPIA-14 [98] | 1539 × 1376, 1663 x 1485 | 4240 | H&E | breast cancer | coordinates of mitosis with a confidence degree / | mitosis detection, nuclear atypia classification |

| | | | | | six criteria to evaluate nuclear atypia | |
|---|---|---|---|---|---|---|
| Kumar et al [99,100] | 1000 x 1000 | 30 | H&E | various cancer | coordinates of annotated nuclear boundaries | nuclear segmentation |
| MITOS 2012 [20,101] | 2084 x 2084, 2252 x 2250 | 100 | H&E | breast cancer | coordinates of mitosis | mitosis detection |
| Janowczyk et al. [102,103] | 1388 x 1040 | 374 | H&E | lymphoma | none | disease classification |
| Janowczyk et al. [102,103] | 2000 x 2000 | 311 | H&E | breast cancer | coordinates of mitosis | mitosis detection |
| Janowczyk et al. [102,103] | 100 x 100 | 100 | H&E | breast cancer | coordinates of lymphocyte | lymphocyte detection |
| Janowczyk et al. [102,103] | 1000 x 1000 | 42 | H&E | breast cancer | mask for epithelium | epithelium segmentation |
| Janowczyk et al. [102,103] | 2000 x 2000 | 143 | H&E | breast cancer | mask for nuclei | nuclear segmentation |
| Janowczyk et al. [102,103] | 775 x 522 | 85 | H&E | colorectal cancer | mask for gland area | gland segmentation |
| Janowczyk et al. [102,103] | 50 x 50 | 277524 | H&E | breast cancer | none | tumor detection |
| Gertych et al[22] | 1200 x 1200 | 210 | H&E | prostate cancer | mask for gland area | gland segmentation |
| Ma et al[104] | 1040x1392 | 81 | IHC | breast cancer | | TIL analysis |
| Linder et al. [64,105] | 93-2372 x 94-2373 | 1377 | IHC | colorectal cancer | mask for epithelium and stroma | segmentation of epithelium and stroma |
| Xu et al. [55] | various size | 717 | H&E | colon cancer | | |
| Xu et al. [55] | 1280 x 800 | 300 | H&E | colon cancer | mask for colon cancer | segmentation |

4.2.1. Efficient labeling

One way to increase training data is to reduce the working time of pathologists to specify ROIs in the WSI. Easy-to-use GUI tools helps pathologists efficiently label more samples in shorter periods of time [42,43]. For example, Cytomine [42] not only allows pathologists to surround ROIs in WSIs with ellipses, rectangles, polygons or freehand drawings, but also applies content-based image retrieval algorithms to speed up annotation. Another interesting idea to reduce working time is to automatically localize ROIs during diagnosis, which uses the usual working time for diagnosis as labeling by tracking pathologists' behavior. This approach tracks pathologists' eye movement [44], mouse cursor positions [45] and change in viewport [46]. However, localizing ROIs accurately from these tracking data is not always easy since pathologist's do not always spend time looking at ROIs, and boundary information obtained by these approaches tends to be less clear.

Another approach that utilizes a machine learning method is active learning [43,47–49,106,107]. This is generally effective when the acquisition cost of label data is large (i.e. pathological images). Active learning is a method used in supervised learning, and it automatically chooses the most valuable unlabeled sample (i.e. the one that is expected to improve the identification performance of classifiers when labeled correctly and used as a training data) and display it for labeling by pathologists. Since this approach is likely to increase discrimination performance with smaller number of labeled images, the total labeling time to obtain the same discrimination performance will be shortened [47]. Many criteria such as uncertainty sampling [43], Query-by-Committee [47], variance reduction [48], and hypothesis space reduction [49] have been proposed for selecting valuable unlabeled samples.

4.2.2. Incorporating insufficient label

Even if the exact position of the ROI in a WSI is not known, it is possible that the information regarding the presence/absence of the ROI in the WSI is available from the pathological diagnosis assigned to the WSI or WSI-level labels. These so-called weak labels are easy to obtain compared to patch-level labels even when the WSIs have no further information, and in this regard, WSIs is considered as a "bag" made with many patches (instances) in machine learning settings. When diagnosing cancer, WSI is labeled as cancer if at least one patch contains cancerous tissue, or normal if none of the patches contain cancerous tissue. This setting is a problem of multiple instance learning [51,108] or weakly-supervised learning [50,52]. In a typical multiple instance learning problem, positive bags contain at least one positive instance and negative bags do not contain any positive instances. The aim of multiple instance learning is to predict bag or instance label based on training data that contains only bag labels. Various methods in multiple instance learning have been applied to histopathological image analysis including boosting-based approach [50], support vector machine-based approach [53] and deep learning-based approach [52].

In contrast, semi-supervised learning [30,54,109,110] utilizes both labeled and unlabeled data. Unlabeled data is used to estimate the true distribution of labeled data. For example, as shown in Figure 1, decision boundary which takes only the labeled samples into account would form a vertical line, but that considering both labeled and unlabeled samples would form a slanting line, which could be more accurate. Since semi-supervised learning is considered particularly effective when samples in the same class form a well-discriminative cluster, relatively easy problem could be a good target.

4.2.3. Reusing parameters from another task

Performing supervised learning using too few training data would only result in insufficient generalization performance. This is true especially in deep learning, where the number of parameters to be learned is very large. In such a case, instead of learning the entire model from scratch, learning often starts by using (a part of) parameters of a pre-trained model optimized in another similar task. Such a learning method is called transfer learning. In CNN, layers before the last (typically three) fully-connected layers are regarded as feature extractors. The fully-connected layers are often replaced by a new network suitable for the target task. The parameters in earlier layers can be used as is [55], or as initial parameters and then the network is learned partially or fully from the training data of the target task [16,56,57] (so-called fine-tuning). In pathological images, no network learned from tasks using other pathological images are available, and thus networks learned using ImageNet, which is a database containing vast number of general images, are often used [16,55–57]. For example, Xu et al., performed classification and segmentation tasks on brain and colon pathological images using features extracted from CNN trained on ImageNet, and achieved state-of-the-art performance [55]. Although the pathological image itself looks very different to the general images (e.g. cats and dogs), they share common basic image structures such as lines and arcs. Since earlier layers in deep learning capture these basic image structures, such pre-trained models using general images work well in histopathological image analysis. Nevertheless, if models pre-trained on sufficient number of diverse tissue pathology images are available, they may outperform the ImageNet pre-trained models.

4.3. Different levels of magnification result in different levels of information

Tissues are usually composed of cells, and different tissues show distinct cellular features. Information regarding cell shape is well captured in high-power field microscopic images, but structural information such as a glandular structure made of many cells are better captured in a lower-power field (Figure 2). Because cancerous tissues have both cellular and structural atypia, images taken at multiple magnifications would each contain important information. Pathologists diagnose diseases by acquiring different kinds of information from the cellular level to the tissue level by changing magnifications of a microscope. Even in machine learning, researches utilizing images at different magnifications exist [58–60]. As mentioned above, it is difficult to handle the images at its original resolution directly, images are often resized to correspond to various magnifications and used as input for analysis. Regarding diagnosis, the most informative magnification is still controversial [14,40,111], but improvement in accuracy is sometimes achieved by inputting both high and low magnification images simultaneously, probably depending on the types of diseases and tissues, and machine learning algorithms.

4.4. WSI as orderless texture-like image

Pathological image is different from cats and dogs in nature, in a sense that it shows repetitive pattern of minimum components (usually cells). Therefore, it is rather closer to texture than object. CNN acquires shift invariance to a certain extent by pooling operations. In addition, even normal CNN can learn texture-like structure by data augmentation by shifting the tissue image with a small stride. Meanwhile, there has been methods which utilize texture structure more intensively, such as gray level co-occurrence matrix [112], local binary pattern [113], Gabor filter bank, and recently developed deep texture representations using a CNN [65,114]. Deep texture representations are computed using a correlation matrix of feature maps in a CNN layer.

Converting the CNN features to texture representations would lead to the acquisition of invariance regarding cell position, while utilizing good representations learned by CNN. Another advantage of deep texture representation is that there are no constraints on the size of input image, which is very suitable for large image size of WSI. The boundary between texture and non-texture is unclear, but a single cell or a single structure is obviously not a texture. Better approach would thus depend on the object to be analyzed.

### 4.5. Color variation and artifacts

WSIs are created through multiple processes: pathology specimens are sliced and placed on a slide glass, stained with hematoxylin and eosin, and then scanned. At each step undesirable effects, which are unrelated to the underlying biological factors, could be introduced. For example, when tissue slices are being placed onto the slides, they may be bent and wrinkled; dust may contaminate the slides during scanning; blur attributable to different thickness of tissue sections may occur (Figure 3); and sometimes tissue regions are marked by color markers. Since these artifacts could adversely affect the interpretation, specific algorithms to detect artifacts such as blur [72] and tissue-folds [74] have been proposed. Such algorithms may be used for preprocessing WSIs.

Another serious artifact is color variation as shown in Figure 4. The sources of variation include different lots or manufacturers of staining reagents, thickness of tissue sections, staining conditions and scanner models. Learning without considering the color variation could worsen the performance of machine learning algorithm. If sufficient data on every stained tissue acquired by every scanner can be incorporated, the influence of color variation on classification accuracy may become negligible; however, that seems very unlikely at the moment.

To address this issue, various methods have been proposed so far including conversion to gray scale, color normalization [66–69], and color augmentation [70,71]. Conversion to grayscale is the easiest way, but it ignores the important information regarding the color representation used routinely by pathologists. In contrast, color normalization tries to adjust the color values of an image on a pixel-by-pixel basis so that the color distribution of the source image matches to that of a reference image. However, as the components and composition ratios of cells or tissues in target and reference images differ in general, preprocessing such as nuclear detection using a dedicated algorithm to adjust the component is often required. For this reason, color normalization seems to be suitable when WSIs analyzed in the tasks contain, at least partially, similar compositions of cells or tissues.

On the other hand, color augmentation is a kind of data augmentation performed by applying random hue, saturation, brightness, and contrast. The advantage of color augmentation lies in the easy implementation regardless of the object being analyzed. Color augmentation seems to be suitable for WSIs with smaller color variation, since excessive color change in color augmentation could lead to the loss of color information in the final classifier. As color normalization and color augmentation could be complementary, combination of both approaches may be better.

### 5. Summary and Outlook

Digital histopathological image recognition is a very suitable problem for machine learning since the images themselves contain information sufficient for diagnosis. In this review, we brought up problems in digital histopathological image analysis using machine learning. Due to great efforts made so far, these problems are becoming tractable, but there is still room for improvement. Most of these problems are likely to be solved once a large number of well-annotated WSIs become available. Gathering WSIs from various institutes to collaboratively annotate them with the same criteria and making these data public will be sufficient to boost the development of more sophisticated digital histopathological image analysis.

Finally, we suggest some potential future research topics that have not been well studied so far.

*Discovery of novel objects*

In actual diagnostic situations, unexpected objects such as aberrant organization, rare tumor (thus not included in training data) and foreign bodies could exist. However, discrimination model including Convolutional Neural Networks forcibly categorizes such objects into one of the pre-defined categories. To solve the problem, outlier detection algorithms, such as one-class kernel principal component analysis[115], have been applied to the digital pathological images but only a few researches have addressed the problem so far. More recently, some deep learning-based methods utilizing reconstruction error [116] have been proposed for outlier detection in other domains, but they are not yet applied in the histopathological image analysis.

*Interpretable deep learning model*

Deep learning is often criticized because its decision-making process is not understandable to humans and therefore often described as being a black box. Although decision-making process of human is not a complete white box either, people want to know the decision process or decision basis. This could lead to a new discovery in the pathology field. Although this problem has not been completely solved so far, some research has attempted to provide solutions, such as joint learning of pathological images and its diagnostic reports integrated with attention mechanism[117]. In other domains, decision basis can be inferredindirectly represented by visualizing the response of a deep neural network[117,118], or presenting the most helpful training image using influence functions[119].

*Intraoperative diagnosis*

Pathological diagnosis during surgery influences intraoperative decision making, and thus could be another important application in histopathological image analysis. As diagnostic time in intraoperative diagnosis is very limited, rapid classification while keeping accuracy is of importance. Due to the time constraint, rapid frozen section is used instead of Formalin-fixed paraffin-embedded (FFPE) section which takes longer time to prepare. Therefore, for this purpose training of classifiers should be performed using frozen section slides. Few research has analyzed frozen sections [120] so far partly because the number of WSIs suitable for the analysis is not sufficient, and task is more challenging compared to FFPE slides.

*Tumor infiltrating immune cell analysis*

Because of the success of tumor immunotherapy, especially immune-checkpoint blockade therapies including

anti-PD-1 and anti-CTLA-4 antibodies, immune cells in tumor microenvironment have gained substantial attention in recent years. Therefore, quantitative analysis of tumor infiltrating immune cells in slides using machine learning techniques will be one of the emerging themes in digital histopathological image analysis. Tasks related to this analysis include detection of immune cells from H&E stained image[121,122] and detection of more specific type of immune cells using immunohistochemistry[104]. Additionally, the pattern of immune cell infiltration and proximity of each immune cells are reportedly related to cancer prognosis[123], analysis of spatial relationships between tumor cells and immune cells, and the relationships between these data and prognosis or response to immunotherapy using specialized algorithms such as graph-based algorithms [63,124] will also be of great importance.

Acknowledgement

This study was supported by JSPS Grant-in-Aid for Scientific Research (A), No. 25710020 (SI).

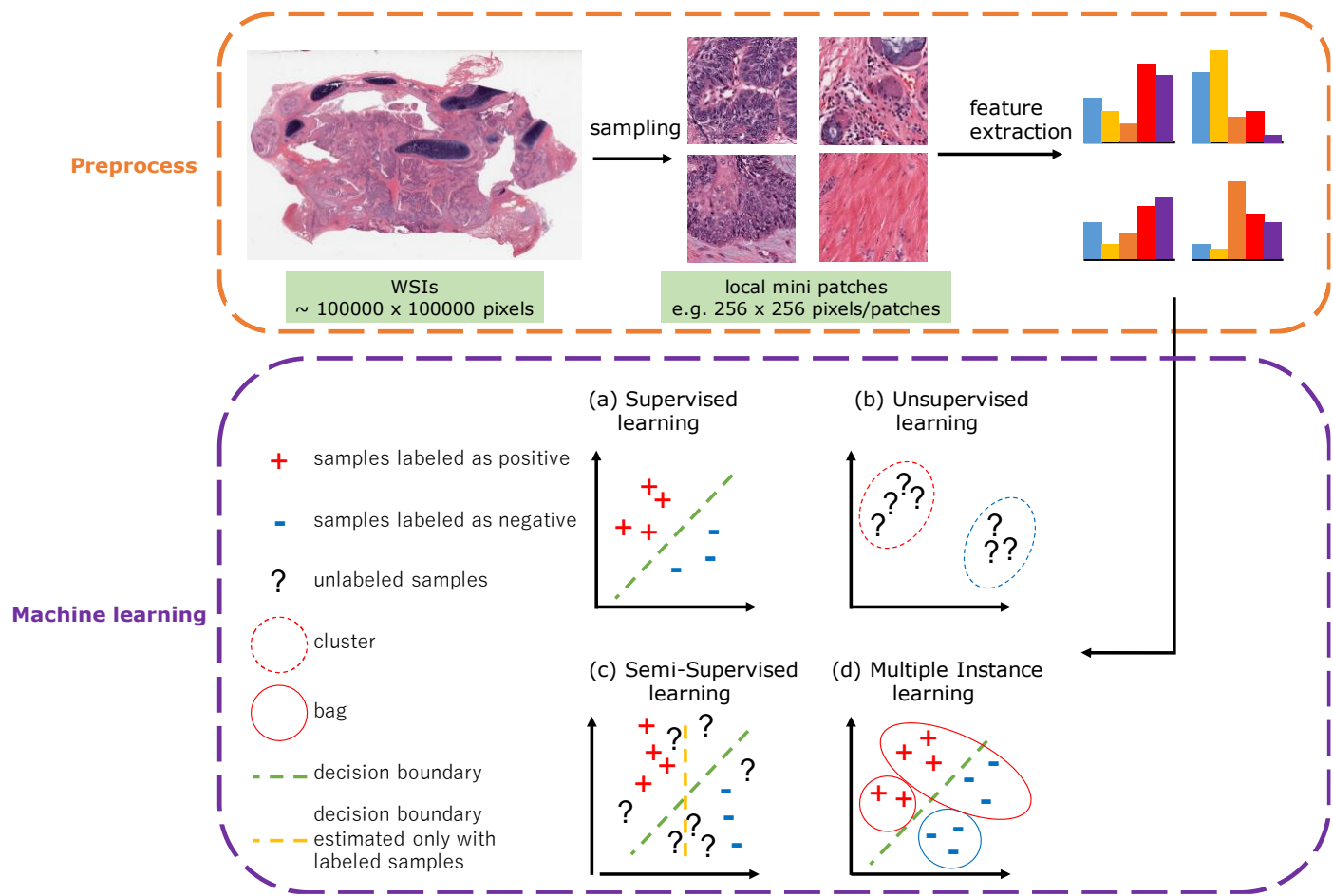

**Figure 1. Typical steps for machine learning in digital pathological image analysis.** After preprocessing whole slide images, various types of machine learning algorithms could be applied including (a) supervised learning (see Section 2), (b) unsupervised learning (see Section 2), (c) semi-supervised learning (see Section 4.2.2), and (d) multiple instance learning (see Section 4.2.2). The histopathological images are adopted from The Cancer Genome Atlas (TCGA)[34]

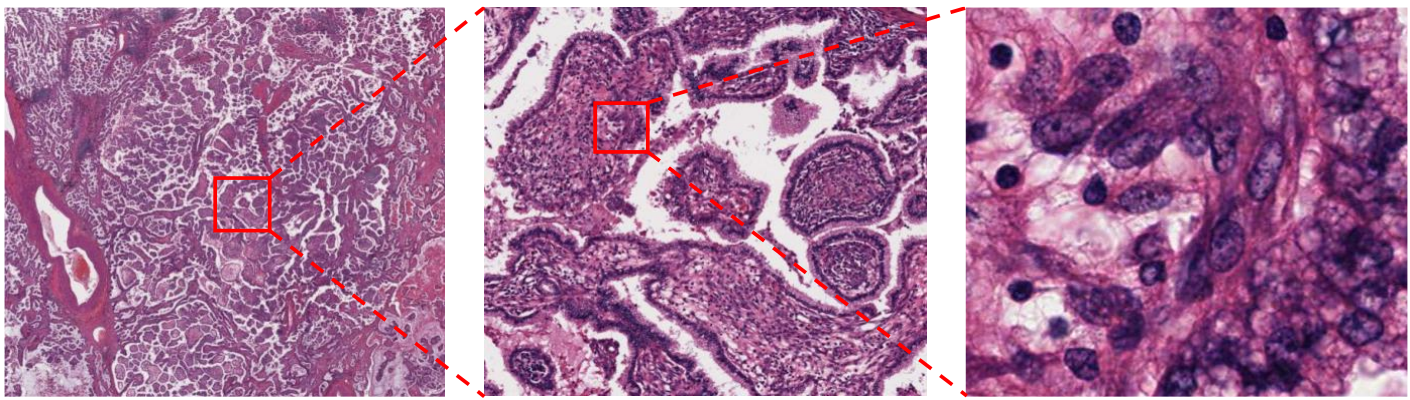

**Figure 2. Multiple magnification levels of the same histopathological image.** Right images show the magnified region indicated by red box on the left images. Leftmost image clearly shows papillary structure, and rightmost image clearly shows nucleus of each cell. The histopathological images are adopted from TCGA[34]

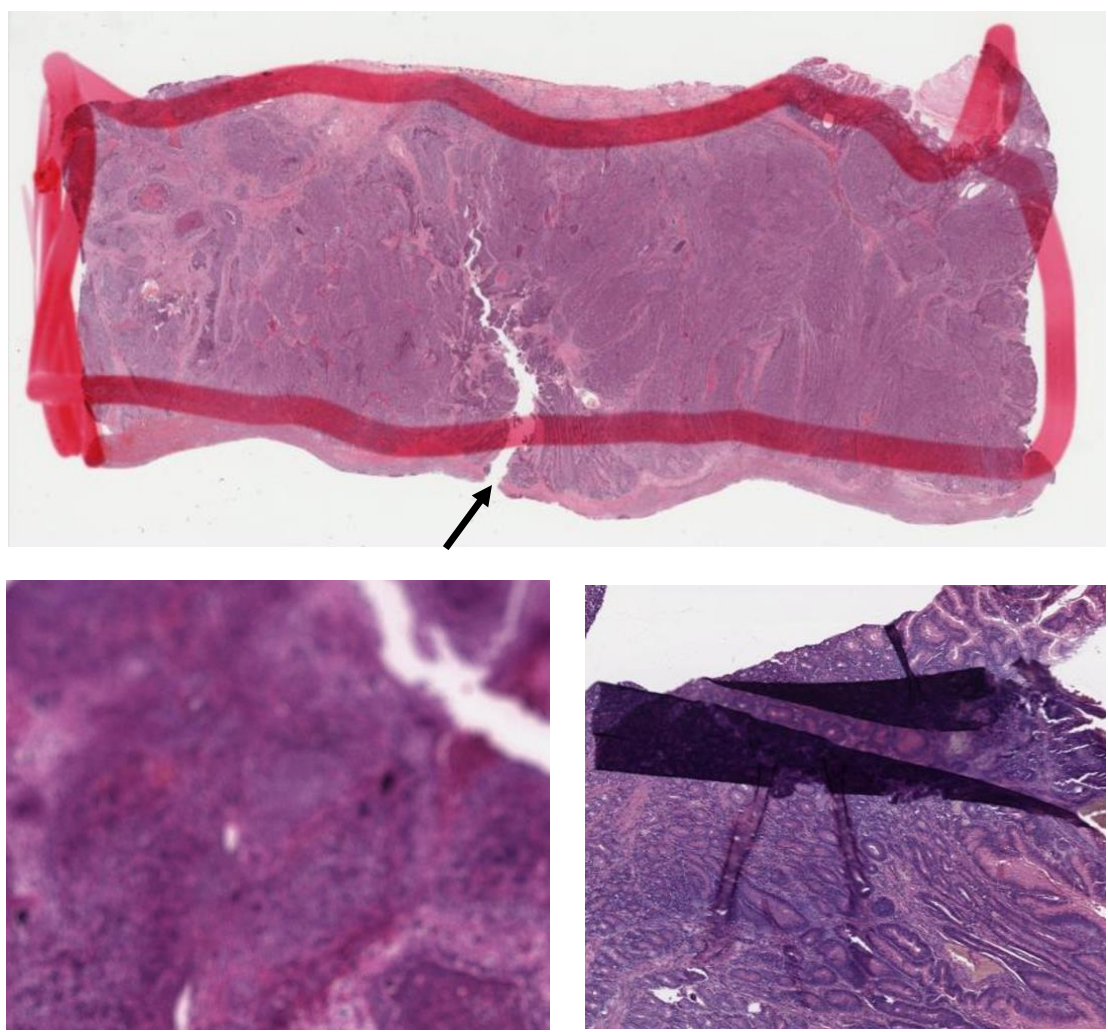

**Figure 3. Artifacts in WSIs.** Top: tumor region is outlined with red marker. The arrow indicates a tear possibly formed during the tissue preparation process. Left bottom: blurred image. Right bottom: folded tissue section. The histopathological images are adopted from TCGA[34]

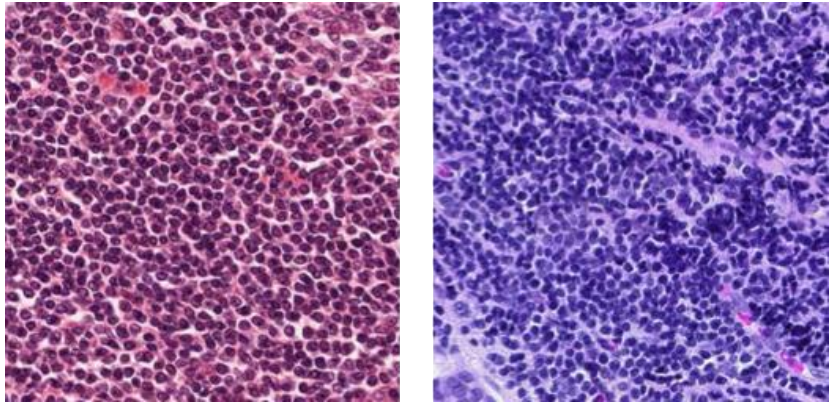

**Figure 4. Color variation of histopathological images.** Both of these two images show lymphocytes. The histopathological images are adopted from TCGA[34]